\def\year{2021}\relax

\documentclass[letterpaper]{article} 
\usepackage{aaai21}  
\usepackage{times}  
\usepackage{helvet} 
\usepackage{courier}  
\usepackage[hyphens]{url}  
\usepackage{graphicx} 
\usepackage{natbib}  
\usepackage{caption} 
\usepackage{ulem}
\usepackage{multirow}
\usepackage{xcolor} 

\title{What Truly Matters? Using Linguistic Cues for Analyzing the \texttt{{\#BlackLivesMatter}} Movement and its Counter Protests: 2013 to 2020}
\author{
    Jamell Dacon\textsuperscript{\rm 1}\thanks{Corresponding Author}, Jiliang Tang\textsuperscript{\rm 1}
    \\
}
\affiliations{
    \textsuperscript{\rm 1}Department of Computer Science, Michigan State University, USA\\

    \{daconjam, tangjili\}@msu.edu 

}

\begin{document}

\maketitle
\begin{abstract}
Since the fatal shooting of 17-year old Black teenager Trayvon Martin in February 2012 by a White neighborhood watchman, George Zimmerman in Sanford, Florida, there has been a significant increase in digital activism addressing police-brutality related and racially-motivated incidents in the United States. 
In this work, we administer an innovative study of digital activism by exploiting social media as an authoritative tool to examine and analyze the linguistic cues and thematic relationships in these three mediums. We conduct a multi-level text analysis on 36,984,559 tweets to investigate users' behaviors to examine the language used and understand the impact of digital activism on social media within each social movement on a sentence-level, word-level, and topic-level. Our results show that excessive use of racially-related or prejudicial hashtags were used by the counter protests which portray potential discriminatory tendencies. Consequently, our findings highlight that social activism done by Black Lives Matter activists does not diverge from the social issues and topics involving police-brutality related and racially-motivated killings of Black individuals due to the shape of its topical graph that topics and conversations encircling the largest component directly relate to the topic of \textit{Black Lives Matter}. Finally, we see that both Blue Lives Matter and All Lives Matter movements depict a different directive, as the topics of \textit{Blue Lives Matter} or \textit{All Lives Matter} do not reside in the center. These findings suggest that topics and conversations within each social movement are skewed, random or possessed racially-related undertones, and thus, deviating from the prominent social injustice issues.

\end{abstract}

\section{Introduction}\label{sec:intro}
Over the past decade there has been an uprising in social (public-affairs) movements due to an increase in police-brutality related and racially-motivated cases across the United States of America. Indeed, there has been an drastic increase in offline and digital (online) activism where citizens have shown their uneasiness in several ways such as protesting, educating, demanding social justice, looting and rioting \cite{Garza:2014, Reissner2019FromT, Jackson2020OnA}. With the growth of social media platforms such as Facebook, Instagram, Twitter, Reddit, etc., racial and discriminatory incidents are now recorded, documented and shared via social platforms in real time to showcase these said incidents. These incidents are shown and shared from a raw perspective compared to lucid conventions practiced by conventional media outlets (e.g newspapers) when approaching social and political events and addressing social injustice issues. Therefore, social media plays a key role in sharing of critical information of police-brutality related and racially-motivated incidents that the general public may have been unaware of \cite{Zulli2020EvaluatingHA, Yang2016NarrativeAI}. In other words, social media has become a definitive tool in linking and mobilizing events and people, hence, establishing decisive digital activism via social media.

Specifically, digital activism has been proven to be a useful tactic for mass political mobilization and has opened up new opportunities for protesters to participate. Activists employ hashtag activism accompanied by their thoughts, feelings, opinions, and emotions attempting to address social injustice issues that need immediate attention along with one or more of these three hashtags, i.e., \texttt{{\#BlackLivesMatter}}, \texttt{{\#BlueLivesMatter}}, and \texttt{{\#AllLivesMatter}}, respectively. Nonetheless, there are individuals whose online behaviors include ``ambivalent'' or ``pessimistic'' tendencies focused towards the goals of a particular social movement. For instance, those opposed to or criticizing the Black Lives Matter (BLM) movement used the hashtags \texttt{{\#BlueLivesMatter}} and \texttt{{\#AllLivesMatter}} to (i) \textit{dismiss distinctive police-brutality related cases against Black Americans often insinuating those Black individuals were not paying attention to the significance of police protection}; and (ii) \textit{to redirect attention from peculiar racially-motivated incidents} \cite{doi:10.1177/0160597616643868}. 

In particular, individuals often construe the phrase ``\textit{Black Lives Matter}'' to be divisive of other races and/or ethnicities, hence the creation of the All Lives Matter (ALM) movement which coherently argue that all lives are equal. However, ALM is seen as a direct push against the BLM movement often dismissing, ignoring or denying these police-brutality related and racially-motivated incidents against Black individuals. Similarly, the Blue Lives Matter (BlLM) movement's flag is often flown alongside Confederate Flags by white supremacists wearing former President Donald Trump's ``MAGA'' attire. Despite BlLM being heavily associated with racist supporters BlLM activists generally assert professional pride. Nevertheless, BLM has gained momentous international media and political attention following the inequitable high-profile police-brutality related and racially-motivated killings of Ahmaud Arbery, Breonna Taylor and George Floyd in 2020, and continues to be a representation of an ongoing social (`modern day' civil and human rights) movement that attests for justice against racism, and protests against police-brutality and racial-inequality towards Black people. 

In this work, we administer an innovative linguistic and practical study of digital activism within the three large-scale social movements by exploiting tweets as an authoritative tool to examine and analyze the linguistic cues (i.e., \textit{thematic relationships and relevance}) in each medium via social media. Our motivation is to assess the behavioral relationships of social activists within each movement, and identify similarities and discrepancies in users' behaviors and topics associated with each social movement keyword, specifically bordering high-profile social injustice issues in the United States. The main goal of our study is to understand the impact of digital activism within each social movement to determine if a user's behavior is ``optimistic'', ``ambivalent'' or ``pessimistic'' towards social issues. Our main tasks are to investigate each medium on sentence-level, topic-level, and word-level, respectively and their respective hashtags, namely \texttt{\#BlackLivesMatter}, \texttt{\#BlackLivesMatter} and \texttt{\#AllLivesMatter}. To achieve this goal, we collected approximately 37 million tweets, to qualitatively examine hashtag activism and global attention, keyword extraction, and discover the thematic relationships and relevance of topics with each medium via thematic text analysis. To do so, we construct two categories of social movement datasets i.e., \textit{100K sampled-datasets} and \textit{1K sampled-datasets}, to visualize and display their thematic relationships and respective topical graphs. To this, we aim to answer the following research questions:
\begin{enumerate}

    \item[--]\textbf{RQ1:} Do these counter protests also increase awareness and global attention or simply incite racism, prejudice and/or hate speech towards the Black community? 

    \item[--]\textbf{RQ2:} Are these counter protests truly addressing pressing social and political issues or easily dismissing, ignoring or denying these police-brutality related and racially-motivated against Black individuals? 
    
    \item[--]\textbf{RQ3:} What are the most central topics and influential tweets of each large-scale social movement; and are they ``optimistic'', ``ambivalent'' or ``pessimistic'' towards pressing social issues?
    
\end{enumerate}

\section{Background and Related Works}\label{sec:related}

In particular, these three large-scale social movements (SMs) address social injustice incidents from different perspectives as follows: (i) Black Lives Matter\footnote{https://blacklivesmatter.com/about/} (BLM) is a non-violent social and political (`modern day' civil and human rights) movement which protests against police-brutality related and racially-motivated incidents i.e. \textit{harassment and killings of black people} \cite{Reissner2019FromT, Garza:2014} and acknowledges that Black and indigenous lives do indeed matter both locally and globally. Today, this SM has been the most influential movement since the civil rights movement; (ii) Blue Lives Matter (BlLM) is a counter protest that mainly attests that killing law enforcement officers should be condemned as a hate crime; and lastly, (iii) All Lives Matter (ALM) is counter protests against BLM which advocates a skewed version of \textit{racial-equality} from a ``race-neutral'' perspective to address controversial social and political incidents. However, one frequent criticism is that ALM attempts to mitigate and suppress high-profile police-brutality related killings and racially-motivated incidents towards to Black lives in the United States by redirecting attention, which is seen as a form of prejudice or racism towards Blacks individuals \cite{Mundt2018ScalingSM}.

As a result of the popularity and boundless capabilities of social media, there have been a significant increase in online activism, in which SMs gain and maintain vast attention in order to achieve their goals. In \cite{KEIB2018106}, the authors focus on four aspects of importance (1) Policy or Action, (2) Group, (3) Social Actor and Politics, and (4) Sentiment direction as social media connects content to the public via high quality content. Findings suggest that tweets that were more likely to be retweeted were of strong importance or were associated with policy or action. The authors notice that neutral tweets are less likely to be retweeted compared to emotional tweets, and thus, revealing an increase effectiveness for social movements.

Researchers have studied the discourse in protest and counter protests from the perspective of race theory and critical theory as Twitter is a large platform for sociological analysis \cite{doi:10.1177/0160597616643868, gallagher2017divergent, blevins2019tweeting}. In \cite{gallagher2017divergent}, the authors conduct a multi-level analysis of 860,000 tweets to demonstrate the key differences between the protest groups, namely, ALM and BLM via word-level analysis and topic level analysis. By adopting approaches in previous works on political polarization they use hashtags as a proxy for topics and analyze the dynamics of the usage of words in tweets on a micro-level within the \texttt{\#BlackLivesMatter} and \texttt{\#AllLivesMatter} hashtags. Recently, hashtag activism has become very popular on most, if not all social media platforms, where \cite{blevins2019tweeting} focuses on a social network analysis approach to examine prominent hashtags that were used on Twitter following the killing of Mike Brown in Ferguson, Missouri in 2014. 

There has also been several approaches to investigate social media's role in protest movements. In \cite{ince2017response}, the authors focus on BLM's presence and influence of social media by examining how users mesh with the BLM protests via hashtag activism. They propose the notion of ``distributed framing'' regarding the decentralized interactions by analyzing 66,159 \texttt{\#BlackLivesMatter} tweets from the year 2014. They later mention that the hashtags associated with BLM promote unity and advocate for the movement while addressing the pressing issues of police-related brutality and racially-motivated killings. In addition, some works attempt to evaluate large-scale social movements by exploring the role of social media in aiding in the expansion, and overall strengthening of the movement \cite{Mundt2018ScalingSM}.
\section{Methodology}\label{sec:methods}
\begin{table*}[t]
\centering
\scalebox{0.9}{\begin{tabular}{lccccc} 
\hline
  & \textbf{Tweets} & \textbf{Users} & \textbf{Retweets} & \textbf{Top Languages}  \\ 
\hline
\textbf{All}   &   36,984,559  & 7,010,043  & 20,748,235  &  en, fr, ja, es, pt, th, de, it, nl, ko  \\
\textit{BlackLivesMatter} &  29,173,435 &  5,665,400  & 17,260,142  &  en, fr, ja, es, pt, th, de, it, nl, ko \\
\textit{AllLivesMatter}   &  2,088,326  &  721,741  &   1,349,035  &  en, es, nl, ja, fr, de, pt, it, tl, ht \\
\textit{BlueLivesMatter}  &    2,830,532  &  418,485 &  1,491,035 &  en, es, fr, pt, nl, it, ja, de, ht, tl \\
\hline
\end{tabular}}
\caption{Descriptive counts for our collected dataset and each keyword, respectively. ISO 639-1 Language codes: en = English, fr = French, ja = Japanese, es = Spanish, pt = Portuguese, th = Thai, de = German, it = Italian, nl = Dutch, ko = Korean, tl = Tagalog, ht = Haitian, Haitian Creole. Note that tweets may contain one or more of the keywords and thus, being added to multiple rows.}\label{tab:data}
\end{table*}

\begin{figure*}[t]
    \includegraphics[scale=0.55]{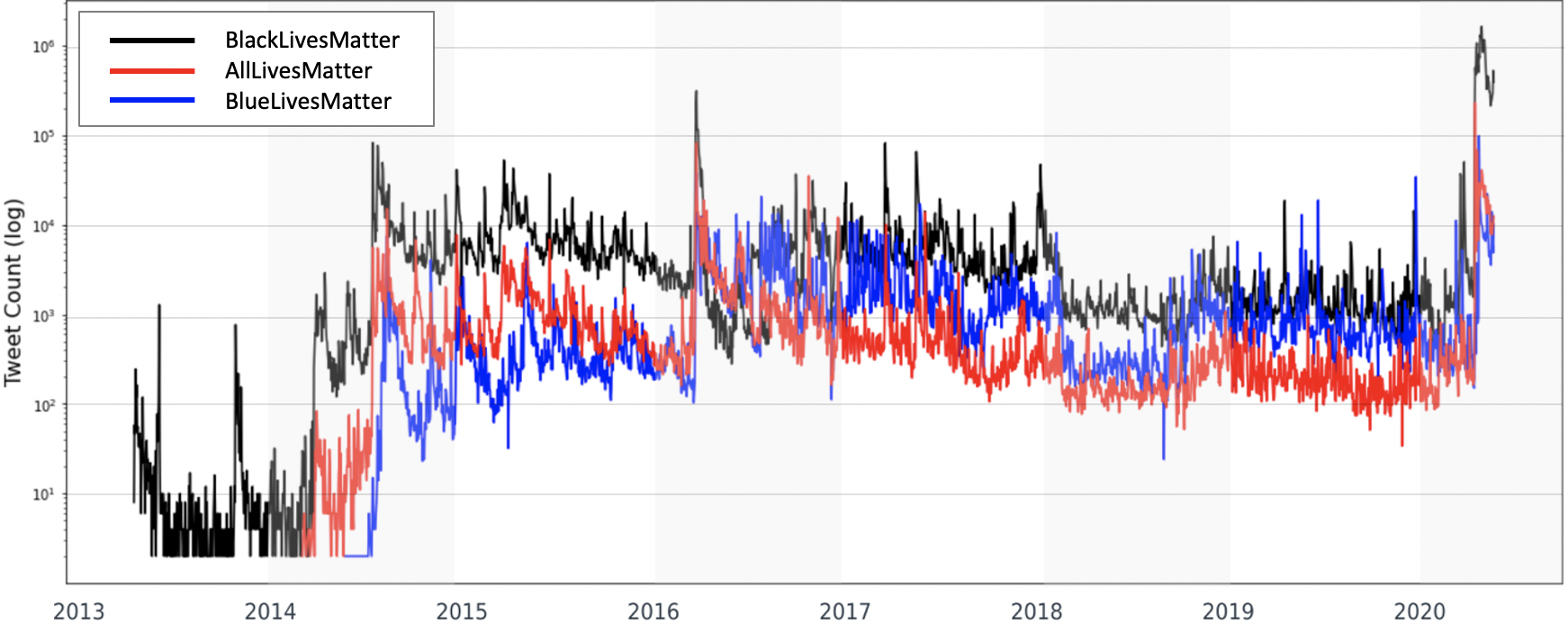}
    \centering
    \caption{Logged daily frequencies for Tweets of each week of all three keywords spanning from February 6\textsuperscript{\rm th}, 2013 to June 30\textsuperscript{\rm th}, 2020.}
    \label{fig:dates}
\end{figure*}

\subsection{Data Collection}\label{sec:data}
We collect currently available tweets from \cite{giorgi2020twitter}\footnote{Data available at: https://doi.org/10.5281/zenodo.4056563} s.t. our gathered dataset contains IDs of 37 million (36,984,559) tweets containing one or more of the three social/political movements keywords: \textit{BlackLivesMatter}, \textit{AllLivesMatter} and \textit{BlueLivesMatter} from 7 million (7,010,043) users. The dataset possesses (presently) available tweets that span from the beginning of the BLM movement spanning from February 6\textsuperscript{\rm th}, 2013 to June 30\textsuperscript{\rm th}, 2020, a month after the high-profile police-brutality killing of George Floyd. After the original dataset was made publicly available, during the period June 30\textsuperscript{\rm th}, 2020 through December 4\textsuperscript{\rm th}, 2020 (the final day of data collection) tweets, descriptions and user profiles may have been banned, deleted or removed at any point from Twitter by the user or Twitter, or a user's tweets may now be protected. Thus, the numbers reflected in Table~\ref{tab:data} represent currently available tweets and \textit{unique} user profiles in our collected dataset on the final day of collection. Table~\ref{tab:data} also includes the counts for the following: Retweets (Original tweets shared by other users that can also be quoted) and the Top 10 languages of each movement (the language of each tweet automatically detected by Twitter).

We later aggregate tweets with the three respective social/political movements keywords previously mentioned to separate each SM into distinct keyword-datasets to obtain corresponding values such as the number of tweets, users, retweets and top languages which can also be seen in Table~\ref{tab:data}. As tweets can be very noisy having unusual spelling patterns, emojis, pictures, videos, URLs, etc., we randomly sample 12,500 English tweets for each year from 2013 to 2020, resulting in 100,000 tweets from each keyword-dataset i.e., \textit{100KBLM}, \textit{100KBlLM} and \textit{100KALM} datasets to drastically reduce such noise\footnote{Retweets were ignored in the sampling process}.

\subsection{Preprocessing}
Due to the noisiness of tweets, the 100K sampled-datasets were cleaned and processed before beginning any analysis. Specifically, we remove all emojis, URLs, ``@'' handles of users mentioned, and the ``RT'' texts which indicate that a tweet was retweeted by another user. We later remove all punctuation by
implementing a NLTK\footnote{\url{https://www.nltk.org/}} module
that possesses a tokenizer that parts a string utilizing a customary articulation, which coordinates with either the tokens or the separators between tokens, and lowercase the words before removing the stop words to reduce the dimensionality of each tweet. We then filter each tweet through a python package called Word Ninja\footnote{https://github.com/keredson/wordninja} which probabilistically splits connected words utilizing natural language processing (NLP) dependencies from English Wikipedia uni-gram frequencies (e.g., ``\texttt{justiceforgeorgefloyd!!!}'' to ``\texttt{justice for george floyd}''.

\subsection{Keyword Extraction}
Social media has become the most popular means for its users to express their thoughts, feelings, and opinions on any given topic at any give time, thus connecting users via similar interests worldwide. Specifically, Twitter users can produce millions of tweets per day with creates tons of unstructured text data. To break down human language in the form of structured and unstructured text 
data, so it can be comprehended and evaluated by machines, keyword extraction also known as \textit{keyword detection} or \textit{keyword analysis} is a text analysis approach that identifies the most frequently used and significant vocabulary from a text automatically (i.e., \textit{keyword extraction combines both machine learning artificial intelligence (AI) and NLP to pinpoint the usage important words}). It facilitates the succinct summation of textual information extracting the top \textit{n} noun phrases that can predominantly represent a given document and can also identify the central topics presented. 

\subsection{Thematic Analysis}
Thematic Analysis, also known as \textit{topical investigation} is a well-suited subjective strategy that can be utilized when investigating complex and enormous subjective qualitative data (e.g., text, audio recordings, documents, etc.) to identify universal themes, topics, patterns and/or ideas \cite{doi:10.1177/1609406917733847}. It can be argued that thematic analysis is a good approach for assessing diverse users' perspectives thus highlighting similarities and differences, and generating unexpected insights. Thematic analysis may likewise be utilized to sum up significant parts of a major informational collection and exploited to analyze the dynamics of word usage in tweets on a micro-level within each large-scale SM \cite{gallagher2017divergent, doi:10.1177/1609406917733847}.

For further computational analysis, we propose building thematic graph structures that store $\frac{n(n-1)}{n}$ correlation results used to measure how many elements $A_{i,j}$ are in a square matrix, where $i < j$. In particular, the thematic graphs will create clusters of the most-similar topics in the datasets. Visualizing the nodes of the graph with edges above a specific threshold weight \textit{\textbf{w}} will exhibit more structured components (i.e., \textit{relevant topics and common themes of each large-scale SM}). We intend to apply an oratorical strategy by classifying the topics of each graph to determine the most prevalent topics of each SM, and thus, identifying their similarities and differences in discourse on their respective topics. The resulting topical graphs are effective and extremely intuitive as they will contain several larger components (typically at the center being the main theme or topics) with a number of apparent interactions at the edges and a few more moderate areas.

\section{Social Activism via Social Media}\label{sec:activism}
\subsection{Social Justice and Awareness}
Activists for positive change harness and exploit the advantages of social media to spread crucial information, protest, demand social justice and advocate for the goals of a particular movement as social media can quickly connect a single user to a larger audience. For example, Twitter users are capable of easily and quickly sharing information, posting pictures and videos, narrow down their feed with the use of hashtags, and so on. During May, 2020 the BLM movement is an example of a combination of social (i.e., \textit{digital and offline}) activism of millions of citizens across the world who help pressure governments to take action on social and economic issues. In \cite{, 10.1145/3442442.3452325}, the authors note that conventional media outlets for-instance, traditional ``printed'' and online newspapers exhibit several forms of \textit{media biases} such as \textit{selection bias}, \textit{coverage bias}, \textit{presentation bias} and \textit{ideological bias} \cite{Hamborg2018AutomatedIO} by strictly following formal or logical conventions when approaching social and political events \cite{rezapour2018using}. The authors later mention that conventional media outlets continuously display a conscious act of sustaining a systematically biased tendency \cite{Williams}, thus impacting global social awareness and neighboring activists.

\subsection{Social Movements and Hashtags}
As previously mentioned, social media has become a medium and instrument exploited to link and mobilize events and people, as well as maintain digital activism \cite{rezapour2018using}. Several previous studies have identified hashtags as important components of SMs \cite{ince2017response, gallagher2017divergent, 10.1145/2124295.2124320, rezapour2018using} as hashtags are the quickest way to become involved in digital activism unlike traditional media. \textit{Hashtags} aid in the amplification of topics, capture the attention of the younger generation and connects a greater online audience with similar interest and focus. Specifically, the hash (\texttt{\#}) character precedes a string of characters on social media and have been seen as subject identifiers or contextual clues to the tweet's content \cite{10.1145/2124295.2124320}. As a result, social activists systematically leverage hashtags on social media as both unique and universal designators of providing an opportunity to partake in digital activism to address social issues, increase social awareness worldwide, and influencing governments to act expeditiously to focus on said issues. 

In recent years, for example, Twitter users partake in hashtag activism aiding these social issues gaining historic international media and political attention in the United States. By highlighting said issues through their thoughts, feelings, and opinions using one or more of these three popular large-scale SM hashtags, i.e., \texttt{{\#BlackLivesMatter}}, \texttt{{\#BlueLivesMatter}}, and \texttt{{\#AllLivesMatter}}, respectively. The use of the most influential hashtag to date, \texttt{\#BlackLivesMatter} is used to protest against police-brutality related  and racially-motivated related incidents, amplify social awareness of social and political injustice issues surrounding Black lives, yet continuously advocate for unity and contribute to the educational objectives of BLM activists \cite{Freelon2016BeyondTH}. Specifically, its primary use is to address four types of racism: (1) Systemic Racism, (2) Systematic Racism, (3) Structural Racism and (4) Environmental Racism. Following the high-profile police-brutality related killing of George Floyd and Breonna Taylor there have been protests, looting and riots worldwide, as their unjustified and wrongful deaths gained massive global attention. 
\section{Analysis and Results}\label{sec:results}
\textit{\textbf{\underline{Disclaimer:}}} Due to the overall purpose of the study, several terms in the figures and tables may be offensive or disturbing (e.g. prejudicial, racist, fascist, sexist, or homophobic slurs).
These terms are not filtered as they are representative of essential aspects in each dataset.

\begin{table*}[t]
\centering
\scalebox{0.8}{\begin{tabular}{cc|cc|cc}
\hline
\multicolumn{6}{c}{\textbf{Top \textit{k} Hashtags}} \\
\hline
\textbf{BLM} & \textbf{Count} & \textbf{BlLM} & \textbf{Count} & \textbf{ALM} & \textbf{Count}\\
\hline \hline
\texttt{{\#BlackLivesMatter}} & 73,223 & \texttt{{\#BlueLivesMatter}} & 65,260  & \texttt{{\#AllLivesMatter}}   & 43,613  \\
\texttt{{\#blacklivesmatter}} & 9,852 & \texttt{{\#PJNET}} & 12,079  & \texttt{{\#BlackLivesMatter}}  & 18,238  \\
\texttt{{\#BLM}} & 2,475  & \texttt{{\#BackTheBlue}} & 10,871 & \texttt{{\#alllivesmatter}} & 6,751 \\
\texttt{{\#GeorgeFloyd}} & 1,880  & \texttt{{\#BlackLivesMatter}} & 7,468 & \texttt{{\#BlueLivesMatter}} & 3,544 \\
\texttt{{\#AllLivesMatter}} & 1,451 & \texttt{{\#bluelivesmatter}} & 5,452  & \texttt{{\#blacklivesmatter}} & 3,227 \\
\texttt{{\#BLACKLIVESMATTER}} & 1,202  & \texttt{{\#AllLivesMatter}}	& 4,695  & \texttt{{\#RIPGeorgeFloyd}} & 1,319 \\
\texttt{{\#Blacklivesmatter}} & 1,011  & \texttt{{\#ThinBlueLine}}	& 3,387  & \texttt{{\#Minneapolisprotests}}	& 1,302 \\
\texttt{{\#JusticeForGeorgeFloyd}} & 918  &\texttt{{\#MAGA}}	& 2,624 &	\texttt{{\#Amish}}	& 1,300 \\
\texttt{{\#blm}} & 617  & \texttt{{\#PoliceLivesMatter}}	& 2,254 &	\texttt{{\#BLM}}	& 1,139 \\
\texttt{{\#ICantBreathe}} & 586  &\texttt{{\#Police}}	& 1,081 &	\texttt{{\#PJNET}}	& 1,096 \\
\texttt{{\#BTS}} & 558  & \texttt{{\#blacklivesmatter}}	& 1,041 &	\texttt{{\#ALLLIVESMATTER}}	& 790 \\
\texttt{{\#MatchAMillion}} & 508  & \texttt{{\#BLUELIVESMATTER}}	& 962 &	\texttt{{\#WhiteLivesMatter}}	& 613\\
\texttt{{\#Ferguson}} & 497  & \texttt{{\#tcot}}	& 822 &	\texttt{{\#MAGA}} &	603	\\
\texttt{{\#BlueLivesMatter}} & 485  & \texttt{{\#NYPD}}	& 792 &	 \texttt{{\#bluelivesmatter}}	& 580 \\
\texttt{{\#BTSARMY}} & 484  & \texttt{{\#alllivesmatter}}	& 757 &  \texttt{{\#Alllivesmatter}}	& 556\\
\texttt{{\#SayHerName}} & 477 & \texttt{{\#BLM}}	& 756 &	\texttt{{\#blm}}	& 538\\
\texttt{{\#BlackTwitter}} & 417  & \texttt{{\#Antifa}}	& 745 & \texttt{{\#GeorgeFloyd}}	& 521 \\
\texttt{{\#ARMY}} & 379  & \texttt{{\#police}}	& 709 &		\texttt{{\#BackTheBlue}}	& 507 \\
\texttt{{\#dcprotest}} & 363  & \texttt{{\#thinblueline}}	& 666 &	\texttt{{\#AllLivesMatters}}	& 495  \\
\texttt{{\#ArmyMatchedForBLM}} & 339  & \texttt{{\#2A}}	& 629 &	\texttt{{\#BLACKLIVESMATTER}}	& 369\\
\texttt{{\#ARMYMatch1M}} & 330  & \texttt{{\#WhiteLivesMatte}}r	& 561 &	\texttt{{\#Ferguson}}	& 366  \\
\texttt{{\#SandraBland}} & 310  & \texttt{{\#Bluelivesmatter}}	& 526 &	\texttt{{\#cot}} & 362 \\
\texttt{{\#Seattle}} & 300  & \texttt{{\#TCOT}}	& 480 &	\texttt{{\#PoliceLivesMatter}}	& 339\\
\texttt{{\#CHAZ}} & 292  & \texttt{{\#RIP}}	& 474 &	\texttt{{\#MuslimBan}}	& 337 \\
\texttt{{\#racism}} & 287  & \texttt{{\#backtheblue}}	& 466 & \texttt{{\#UnbornLivesMatter}} & 334  \\
\texttt{{\#Antifa}} & 285  & \texttt{{\#Trump}}	& 462 &	\texttt{{\#ProLife}}	& 327\\
\texttt{{\#PhilandoCastile}} & 283  & \texttt{{\#WakeUpAmerica}}	& 443 & \texttt{{\#ThinBlueLine}}	& 321\\
\texttt{{\#EricGarner}} & 281  & \texttt{{\#ccot}}	& 430 &	\texttt{{\#WakeUpAmerica}}	& 311 \\
\texttt{{\#icantbreathe}} & 280  & \texttt{{\#AltLeft}}	& 424 &	\texttt{{\#justice4cephus}}	& 291 \\
\texttt{{\#RIPGeorgeFloyd}} & 271  & \texttt{{\#AntifaTerrorist}}	& 414 &	\texttt{{\#CharlesKinsey}}	& 291 \\
\hline
\end{tabular}}
\caption{Top k hashtags and their respective counts found in each of the three distinct SMs.}\label{tab:hastags}
\end{table*}

\begin{figure*}[!htbp]
    \includegraphics[scale=0.38]{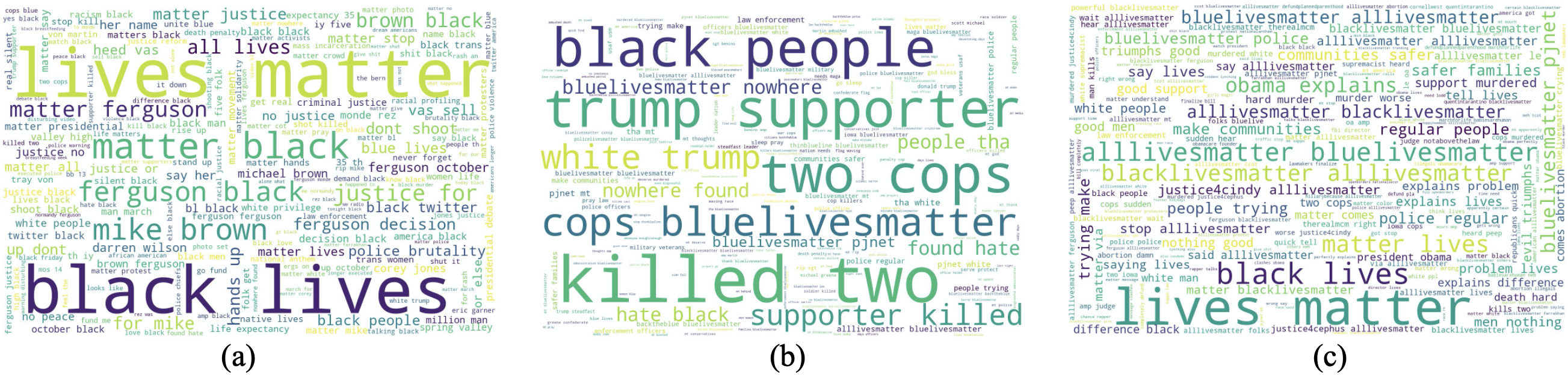}
    \centering
    \caption{Word clouds of (a) BLM, (b) BlLM and (c) ALM illustrating the most frequently occurring bigrams connected by color. Huge printed words directly correlate with the significance of each word.}
    \label{fig:wordcloud}
\end{figure*} 

\begin{figure*}[t]
    \includegraphics[scale=0.7]{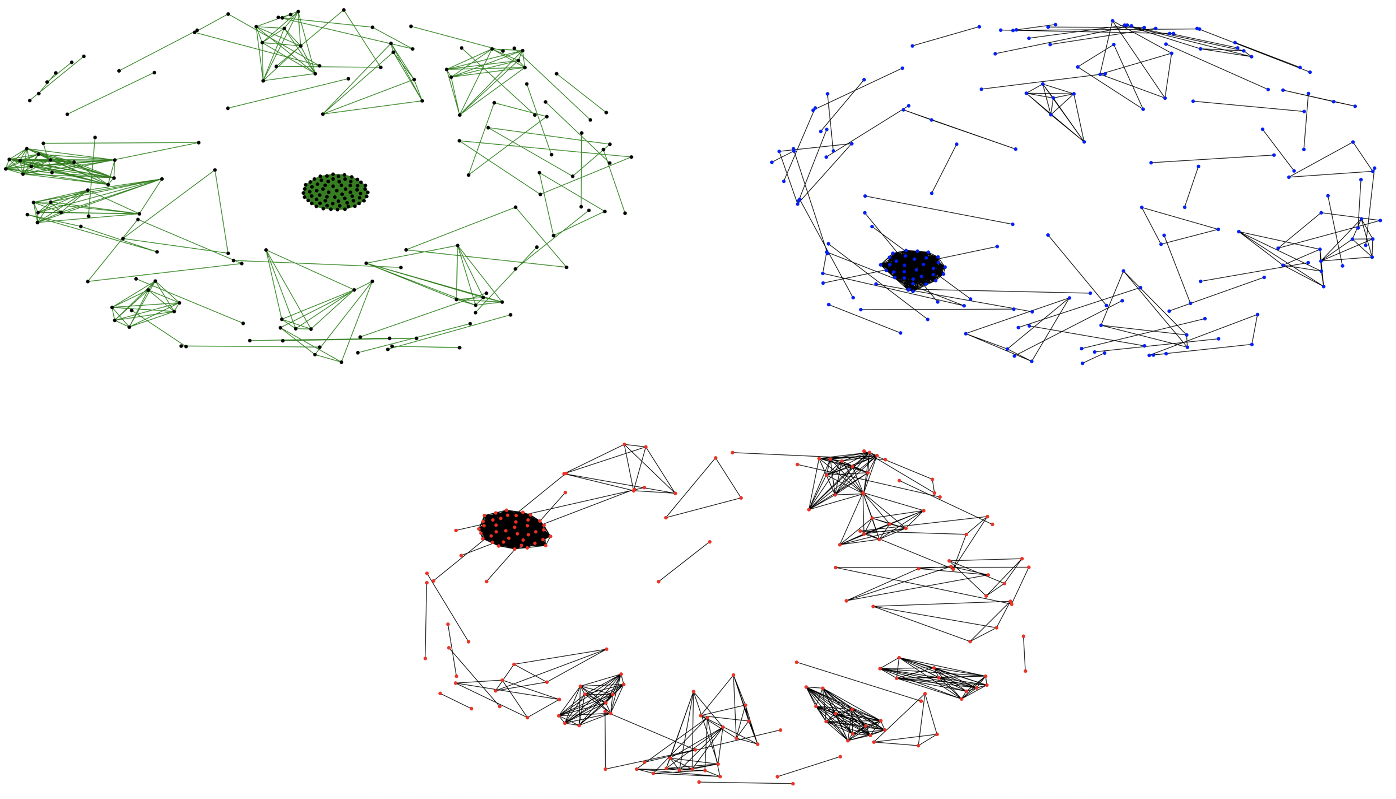}
    \centering
    \caption{Topical graphs via thematic text analysis of each 1K sampled-dataset BLM (top-left) represented by green edges and black nodes, BlLM (top-right) represented by black edges and blue nodes, and ALM (bottom) represented by black edges and red nodes.}
    \label{fig:graphs1}
\end{figure*}

\begin{figure*}[t]
    \includegraphics[scale=0.53]{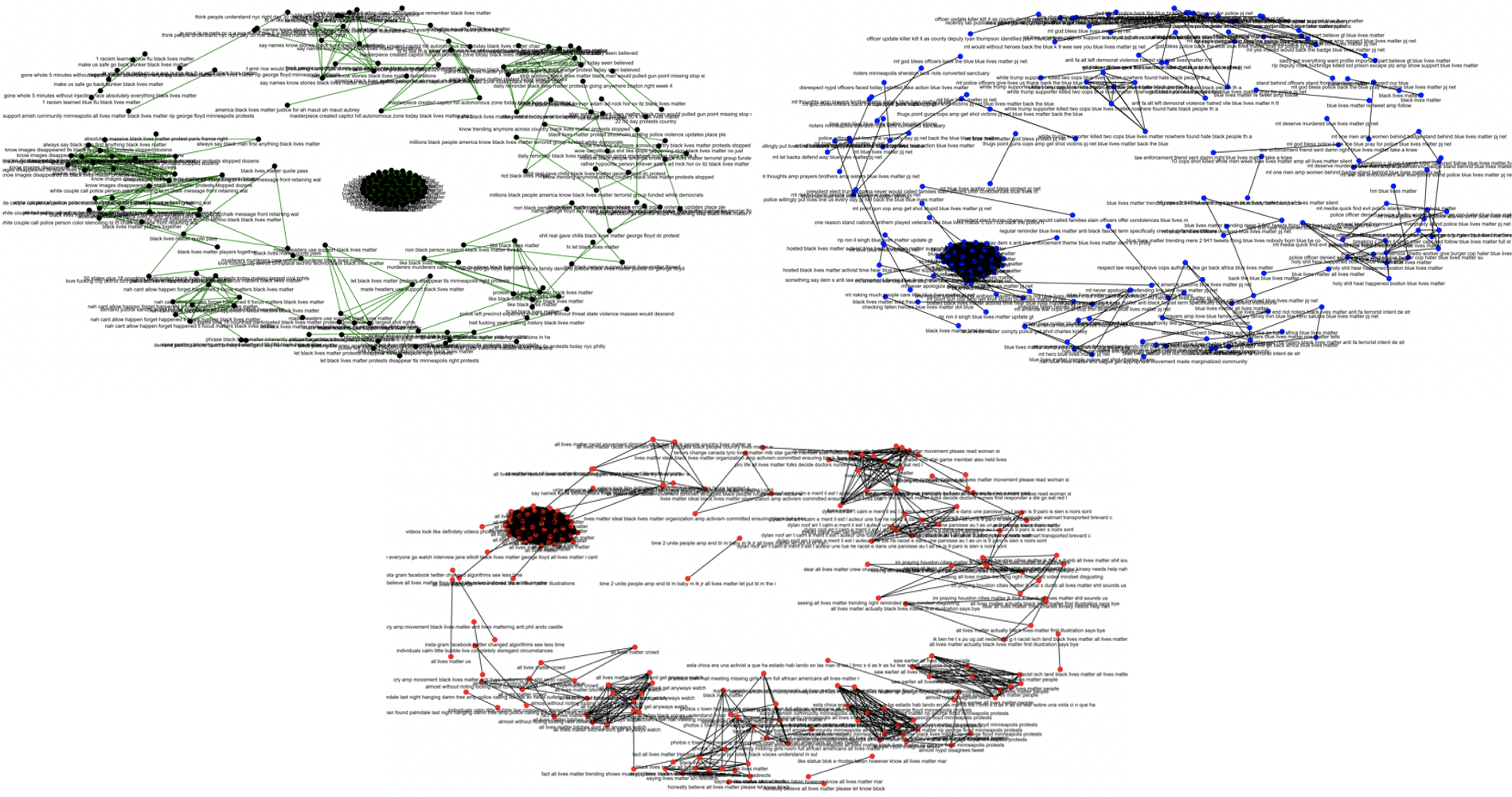}
    \centering
    \caption{Labelled topical graphs via thematic text analysis of each 1K sampled dataset BLM (top-left) represented by green edges and black nodes, BlLM (top-right) represented by black edges and blue nodes, and ALM (bottom) represented by black edges and red nodes. Note that each node is labeled with its corresponding tweet --- which can make tweets in close proximity difficult to read.}
    \label{fig:graphs2}
\end{figure*}

\begin{table*}[t]
\centering
\scalebox{0.8}{\begin{tabular}{c|l|l}
\hline
\multicolumn{1}{c|}{\textbf{Dataset}} & \multicolumn{1}{c|}{\textbf{Tweet Topics}} & \multicolumn{1}{c}{\textbf{Weight}} \\ \hline \hline
1KBLM & \begin{tabular}[c]{@{}l@{}}(i)\texttt{black lives matter}\\ (ii) \texttt{know images disappeared tls black lives matter protests stopped dozens} \\ (iii)\texttt{white couple call police person color stencilling blm chalk message front} \\ \texttt{retaining wal} \\ (iv) \texttt{say names know stories black lives matter illustrations}\\ (v)\texttt{nah cant allow happen forget happened tl focus matters black lives matter}\end{tabular} & \begin{tabular}[c]{@{}l@{}}0.305\\ 0.05\\ \\0.0319\\ 0.0273\\ 0.02728\end{tabular} \\ \hline
1KBlLM & \begin{tabular}[c]{@{}l@{}}(i) \texttt{blue lives matter} \\ (ii) \texttt{white trump supporter killed two cops blue lives matter nowhere found hate} \\\texttt{black people th a} \\ (iii) \texttt{police officer denied service ghetto worker give burger cop hater blue}\\ \texttt{matter su}\\ (iv) \texttt{holy shit hear happened boston blue lives matter}\\ (v) \texttt{respect law respect brave cops authority like go back to africa blue lives matter}\end{tabular} & \begin{tabular}[c]{@{}l@{}}0.161\\ \\0.026\\ \\0.0208\\ 0.0208\\ 0.0156\end{tabular} \\ \hline
1KALM & \begin{tabular}[c]{@{}l@{}}(i)\texttt{all lives matter}\\ (ii) \texttt{support amish community minneapolis all lives matter black lives matter rip} \\ \texttt{george floyd minneapolis protests}\\ (iii) \texttt{lives matter}\\ (iv) \texttt{all lives matter bitches dont get anyways watch}\\ (v) \texttt{saw earlier all lives matter people}\end{tabular} & \begin{tabular}[c]{@{}l@{}}0.206\\ \\ 0.0582\\ 0.0529\\ 0.0476\\ 0.0476\end{tabular} \\ \hline
\end{tabular}}
\caption{Top \textit{k} most central (cleaned and processed) tweet topics illustrating the contents of larger clusters and their corresponding edge weight.}
\label{tab:central}
\end{table*}

\begin{table*}[t]
\centering
\scalebox{0.8}{\begin{tabular}{c|l}
\hline
\multicolumn{1}{c|}{\textbf{Dataset}} & \multicolumn{1}{c}{\textbf{Influential Tweets }}\\ \hline \hline
1KBLM & \begin{tabular}[c]{@{}l@{}}(i) \texttt{black lives matter}\\ (ii) \texttt{know images disappeared tls black lives matter protests stopped dozens}\\ (iii) \texttt{protest end anytime soon black lives matter}\\ (iv) \texttt{made headers use support black lives matter} \\ (v) \texttt{say names know stories black lives matter illustrations}\end{tabular} \\ \hline
1KBlLM & \begin{tabular}[c]{@{}l@{}} (i) \texttt{respect law respect brave cops authority like go back africa blue lives matter}\\ (ii) \texttt{love men blue blue lives matter houston strong} \\ (iii) \texttt{blue lives matter} \\ (iv) \texttt{police officer denied service ghetto worker give burger cop hater blue lives matter su}\\ (v) \texttt{blue lives matter comply police get shot charles kinsey}\end{tabular} \\ \hline
1KALM & \begin{tabular}[c]{@{}l@{}}(i) \texttt{all lives matter}\\ (ii) \texttt{lives matter}\\ (iii) \texttt{urge everyone go watch interview jane elliott black lives matter george lloyd all} \\\texttt{lives matter i cant}\\ (iv) \texttt{all lives matter actually black lives matter first illustration says bye}\\ (v) \texttt{almost nypd disagrees tweet} \end{tabular}\\
\hline
\end{tabular}}
\caption{Top \textit{k} most influential (cleaned and processed) tweets of each topical graph.}
\label{tab:betweenness}
\end{table*}

\subsection{Hashtag Activism and Global Attention}
Since the racially-motivated fatal shooting of 17-year old Black teenager Trayvon Martin to the high-profile police-brutality related killings of Breonna Taylor and George Floyd, and those thereafter, 
hashtag activism has and continues to amplify the marginalized voices of Black individuals. BLM activists frequently attempt to highlight these prejudicial or racist incidents towards Black American lives in the United States by the use of the monumental hashtag \texttt{\#BlackLivesMatter}. To answer RQ1, we intend to examine the use of prominent hashtags as an intermediary to gain a sense of understanding of users' hashtag activism activities. Table~\ref{tab:hastags} displays the top \textit{k} = 30 hashtags of each 100K sampled-dataset. Our findings coincide with the work \cite{ince2017response}, that the most eminent hashtags by BLM activists address the compelling issues encompassing high-profile police-brutality related and racially-motivated incidents while refraining to digress from the goals of the movement like its counter protests. We notice the use of controversial hashtags \texttt{\#WhiteLivesMatter}, \texttt{\#Trump}, \texttt{\#MAGA} and \texttt{\#MuslimBan} used by the counter protests and are often heavily associated with racist, prejudicial or discriminatory tendencies.  

Considering that Twitter has a language detection model which automatically classifies the language of each tweet, we exploit this model to investigate the global attention of each SM and determine the range of digital activism via language. We handle the different languages as a prime tool for approximating social awareness globally and as language is a form of communication that reflects the social uneasiness connecting feelings, thoughts, and emotions. We first create a dictionary to sum the number of languages in our gathered dataset, we then proceed to sum the number of languages found in each keyword-dataset. The 60 ISO 639-1 language codes found in the collected dataset are as follows: en – English, 
fr – French, 
ja – Japanese, 
Es – Spanish, 
pt – Portugese, 
th – Thai, 
de – German, 
it – Italian, 
nl – Dutch/Flemish, 
ko – Korean, 
tr – Turkish, 
ar – Arabic, 
tl – Tagalog, 
ca – Catalan, 
pl – Polish, 
et – Estonian, 
cy – Welsh, 
ru – Russian, 
ht – Haitian/Haitian creole, 
hi – Hindi, 
sv – Swedish, 
da – Danish, 
no – Norwegian, 
fi – Finnish, 
ro – Romanian, 
el – Greek,
lt – Lithuanian,
zh – Chinese,
fa – Persian,
eu – Basque,
vi – Vietnamese,
cs – Czech,
ta – Tamil,
hu – Hungarian,
lv – Latvian,
ur – Urdu,
sl – Slovenian,
is – Icelandic,
mr – Marathi,
bn – Bengali,
uk – Ukrainian,
ml – Malayalam,
ne – Nepali,
si – Sinhala,
kn – Kannada,
bg – Bulgarian,
sr – Serbian,
gu – Gujarati,
pa – Punjabi,
am – Amharic,
ps – Pashto,
or – Oriya,
sd – Sindhi,
hy – Armenian,
te – Telugu,
dv – Divehi,
lo – Lao,
km – Central Khmer,
bo – Tibetan,
ka – Georgian. 
We later learn the number of different languages belonging to each keyword-dataset, i.e., BLM contains all 60 languages, whereas BlLM and ALM contain 42 and 51 different languages, respectfully. 

\subsection{Keyword Extraction}
To answer RQ2, we aim to detect trends of popular thoughts, feelings and opinions of digital activists by investigating keywords and phrases used by each medium. Firstly, we apply \textit{Word Clouds} as word clouds are a simple, yet an efficient information representation strategy for addressing rich textual information in which the size of each word shows its recurrence and/or its significance. In this way, word clouds for each keyword-dataset will consist of main phrases and key expressions without manually having to peruse each line within a given textual dataset. We then add several more common words which were not in the list to further reduce each tweet's dimensionality and avoid frequent impractical words from appearing in each word cloud. These additional words are \textit{dont}, \textit{didnt}, \textit{doesnt}, \textit{cant}, \textit{couldnt}, \textit{couldve}, \textit{im}, \textit{ive}, \textit{isnt}, \textit{theres}, \textit{wasnt}, \textit{wouldnt}, \textit{a}, and \textit{also}. 

We later collect the top \textit{n} = 200 noun phrases that are the most frequently occurring bigrams connected by color for each 100K sampled-dataset. In Figure~\ref{fig:wordcloud} we see the top \textit{n} most frequently used and significant vocabulary in each 100K sampled-datasets of BLM, BlLM and ALM. Huge printed words can give the readers a general understanding of the conversations of each medium. By ignoring the SM keywords (i.e, \textit{BlackLivesMatter}, \textit{BlueLivesMatter} and \textit{AllLivesMatter}) the BLM word cloud contains words such, \textit{mike brown}, \textit{justice for}, \textit{hands up}, \textit{criminal justice}, \textit{ferguson}, \textit{police brutality}, \textit{dont shoot}, \textit{racial profiling}, \textit{white privilege}, \textit{no peace}, etc., directly aligning with addressing social issues surrounding Black lives. In contrast, the BlLM word cloud contains huge printed words such as \textit{black people}, \textit{trump supporter}, \textit{cops}, \textit{killed two}, \textit{white}, \textit{trump}, \textit{hate black}, etc., diverging from the sole topic of professional pride while imitating racially-related and prejudicial undertones. Noun phrases found in the ALM word cloud reference particular components of both BLM and BlLM, yet accompany mentions of the controversial issues of racism, offensive and hate speech in a fascist manner including the opinionated social/political topic of \textit{abortion}.

\subsection{Thematic Text Analysis}
To answer RQ3, we propose to identify universal themes and topics for assessing users' perspectives of tweets belonging to each keyword-dataset. Due to the nature of the dataset, we apply a deductive approach s.t. we assume common themes and/or topics that each 100K dataset should reflect based on prior knowledge and keyword usage as seen in Figure~\ref{fig:wordcloud}. We then apply spaCy\footnote{https://spacy.io/}, an open-source library for advanced NLP tasks such as lemmatization, as lemmatization diminishes curved types of a word. We then utilize a well-known python package called NetworkX\footnote{https://networkx.org} for the analysis and manipulation of our intricate Twitter network structures. After lemmatization and parsing is completed, this library adds an edge if node $i$ is similar to node $j$ s.t. their weight \textit{\textbf{w}} is equal to their similarity result. Note that $i$ and $j$ are tweets of users which belong to the same datasets. To reduce computational complexity, we randomly sample 10\% of each 100K sampled-datasets, collecting 1000 tweets with a random state = 2020, thus creating 1k sampled-datasets (e.g. 1KBLM), resulting in a fully-connected network of 1000 nodes and 499,500 edges created for each 1K sampled-dataset.

Now that we have obtained a fully connected graph, we set the minimum edge weight to 1 (\textit{\textbf{w}} = 1, implying a 100\% correlation of tweets) s.t. we can directly build sparsely-connected topical graph structures creating dense clusters of the most similar tweet topics forming central themes in each 1K sampled-dataset. By setting \textit{\textbf{w}} = 1, each dataset was subjected to deduplication, and thus, each graph retained what we call ``strong edges''. These \textit{strong edges} imply that the remaining nodes have the strongest correlation constituting to larger components and a number of clear connections in the surrounding areas. Subsequently, we remove isolates since each node is heavily correlated with itself. In Figure~\ref{fig:graphs1} we display three distinctive topical graphs by color for each of the 1K sampled-datasets which are a sampled representation of each respective large-scale SMs (i.e. BLM (top-left) represented by green edges and black nodes, BlLM (top-right) represented by black edges and blue nodes, and ALM (bottom) represented by black edges and red nodes). Thoroughly, the resulting force-directed topical graphs contain a compelling number of topics, namely node connected by ``strong edges'', where BLM retained 2540 strong edges, BlLM retained 633 strong edges and ALM retained 1147 strong edges after 100 iterations contributing to the shape of each graph. In other words, nodes with similar scores contribute to a single cluster (i.e., large clusters results from a considerable number of nodes with similar scores, and thus, relating to the same tweet topics (see Figure~\ref{fig:graphs2}).

\subsubsection{Degree Centrality}
Centrality measures are an imperative mechanism for capturing networks as it the simplest form of node connectivity. By manipulating graph theory to calculate the significance of any given node in a network degree centrality discovers relevant parts of the network which inform us about the number of direct `one hop' connections, hence alerting use of the most central topics (or tweets). As seen in Figure~\ref{fig:graphs1}, we present 3 topical graphs via thematic text analysis; however, for a deeper qualitative analysis into the most central themes we attempt to identify and examine the most central tweet topics by displaying the top \textit{k} = 5 largest clusters and their corresponding edge weight to  in Table~\ref{tab:central}.

\subsubsection{Betweenness Centrality}
Betweenness centrality is a pragmatic measure of identifying and calculating the shortest part and determines how often a given node exists on it. Therefore, selecting the nodes that highly influence the structure of the network. Within a network, high betweenness may extensively indicate influential nodes which form disparate clusters. 
For further qualitative analysis, we calculate the betweenness centrality of each of the 1k sampled-datasets to observe and display the top \textit{k} = 5 most influential tweets in the topical graph in Table~\ref{tab:betweenness}.

\section{Discussion}\label{sec:discussion}
As previously mentioned, social media plays a vital role in today's society, especially in regards to documenting events, protesting, and sharing of crucial information of police-related and/or racially-motivated incidents that the general public may have been unaware of \cite{Zulli2020EvaluatingHA, Yang2016NarrativeAI}. In addition, social media has now become a definitive tool for examining and analyzing the relationships between users' online behaviors within each large-scale SM, and identifying the similarities and differences in the discourse on the topics, especially surrounding unjust high-profile police-brutality related and racially-motivated killings of Black individuals in the United States. To raise social awareness insisting on social justice, social activists for positive change mobilize and leverage the advantages of social media to reshape the media biases portrayed to the general public by conventional media outlets.

In this work, we aimed to investigate and analyze the linguistic cues on social media, namely Twitter to examine the language used in the three large-scale SMs, specifically, the Black Lives Matter movement and its counter movements, Blue Lives Matter and All Lives Matter movements. We explored these movements to determine the relationships, and identify similarities and discrepancies in discourse on the topics associated with each keyword (e.g. \textit{BlueLivesMatter}. The primary goal of this work is to understand the impact of digital activism within each social movement on social media. To achieve this goal, we collected approximately 37 million tweets, to qualitatively investigate hashtag activism and global attention, keyword extraction, and observe the thematic relationships and relevance of topics in each medium. Our results show that social activism done by the Black Lives Matter movement does not diverge from the social issues and topics involving police-brutality related and racially-motivated killings of Black individuals as \texttt{black lives matter} is the most prominent topic in the Black Lives Matter topical graph. Due to the shape of the topical graph, this suggests that topics and conversations encircling the largest component directly relate to the topic of \texttt{black lives matter}. 

Conversely, we see that both Blue Lives Matter and All Lives Matter's topical graph depict a different directive, as the largest components of each graph does not reside in the center, but off the the side even after 100 interactions. These findings suggets that topics and conversations within each counter protest are skewed and deviate from the relevant social injustice issues which may be directly related to ``ambivalent'' or ``pessimistic'' user behavior. In addition, both 1KBlLM and 1KALM were subjected to deduplication where they both retained significantly less ``strong edges'' than 1K. This implies that the counter protest's 1K sampled-dataset did not contain a large number of strongly correlated topics but were skewed, random or possessed racially-related or prejudicial undertones, for-instance the excessive use of the hashtags \texttt{\#MAGA}, \texttt{\#WhiteLivesMatter}, \texttt{\#Trump} and \texttt{\#MuslimBan} thus contributing to `ambivalent'' and ``pessimistic'' behavior.

In conclusion, the Black Lives Matter movement is not anti-white or dismissing other races/ ethnicities, nor is it anti-police; however, racial-equality is its goal and police reform is a significant aspect of the movement. Nevertheless, our results highlight social media's influence on social and political issues as keywords act as a catalyst for address said issues that require immediate attention from the public. As there is an increase in attention towards examining and analyzing social movements in the depth, we propose to focus on all three mediums jointly. Being that there is still a need to consider more in-depth qualitative linguistic and practical features, in our future work we intend to examine features such as \textit{emotions}, \textit{language dimensions} and thematic relationships between the three mediums to analyze the joint thematic relevance of particular issues of social injustice. We will aim to further investigate the influence of digital activism and determine the degree of a user's behavior being ``optimistic'', ``ambivalent'' or ``pessimistic'' towards social issues in a collective manner. Therefore, we expect to expand this work to examine the long terms effects on the communities and groups candidly. Thus, engaging in the lived experiences of members of communities directly affected by social activists who simply mention controversial issues in a negatively opinionated (i.e., \textit{racist, prejudicial} or \textit{discriminative}) manner. 

\section{Limitations}\label{sec:limitations}

Unfortunately, there are a few limitations to this study which we are obligated to mention. Firstly, our gathered dataset possesses (presently) available, where we use the term ``presently'' very loosely. Since the last day of data collection, namely December 4\textsuperscript{\rm th}, tweets, profile descriptions that include hashtags, pictures and videos may have been deleted. Additionally, user profiles may also have been banned, deleted, or removed at any point if the user refused to adhere to Twitter's Terms and Conditions from Twitter or by the user if placed under heavy scrutiny, or a user profile may be set to private, and thus, being unable to collect relevant tweets \textit{today}. 

Secondly, we assume that each Twitter account is
operated by a single \textit{unique} user. However, in reality, users can possess multiple accounts for personal and business use. Multiple Twitter accounts can be used for the amplification of social/political issues or conversely, be used to generate noise by diverging conversations and posting racist, fascist, sexist, or homophobic slurs online or simply being a ``troll'' by engaging in prejudicial or discriminatory behaviors. 

Fourthly, although Twitter possesses an automatic language detection model, this model is not entirely accurate as several tweets can easily be misclassified, e.g. ``\texttt{cops no sensible}'' (written in Spanish) is detected as English, and directly translates to ``\texttt{non-sensitive cops}'' when translated to English. We assume that language implies that tweets in the full dataset are from numerous distinct areas worldwide, and not a single or modest number of regions. If geographical location was a \textit{main} experiment in this work, this limitation of geographical misclassification would have severely impacted our findings.

Lastly, to investigate these three large-scale social movements on a sentence-level, topic-level, and a word-level we understand that the sampling of the 3 keyword-datasets was necessary to act as a base for obtaining the 2 sets of sampled datasets, namely, \textit{100K sampled-datasets and 1k sampled-datasets}. Ideally, sampling was done to address the bias issue of the imbalanced gathered dataset proposed by \cite{giorgi2020twitter} as it contained over 75\% of presently available tweets incorporating the keyword \textit{BlackLivesMatter}. In addition, processing and cleaning of tweets for qualitative analysis in hopes to lessen noise by removing emojis, URLs, “@” handles of users, pictures, videos, punctuation, stopwords, lowercasing, and shortening excessive spelling all contribute to the reduction of semantic meanings and emotionality of each tweet. Therefore, emotions of anger, fear, depression, resentment, sadness, anxiety, helplessness and jealousy were moderately impacted.

Furthermore, for further computational analysis, the 1K sampling was necessary to thoroughly reduce computational and time complexity, thus, facilitating the generality in representation. In short, these sampled-datasets are constructed to serve as a generalizable representation of each social movement; however, we must warrant mentioning that these sampled-datasets are not an exact depiction of each movement, yet they do capture important and necessary components of each movement.

\section{Conflicts of Interest and Ethics}\label{sec:conflict}
All authors declare no conflict of interest and that all data used is publicly available and distributed within Twitter’s Terms of Services. The authors state that they also have no known conflicting commercial interests or personal ties that may have impacted the work developed in this paper, or may even be said to have influenced the work presented in this publication.

\bibliography{main}

\begin{thebibliography}{19}
\providecommand{\natexlab}[1]{#1}
\providecommand{\url}[1]{\texttt{#1}}
\providecommand{\urlprefix}{URL }
\expandafter\ifx\csname urlstyle\endcsname\relax
  \providecommand{\doi}[1]{doi:\discretionary{}{}{}#1}\else
  \providecommand{\doi}{doi:\discretionary{}{}{}\begingroup
  \urlstyle{rm}\Url}\fi

\bibitem[{Blevins et~al.(2019)Blevins, Lee, McCabe, and
  Edgerton}]{blevins2019tweeting}
Blevins, J.~L.; Lee, J.~J.; McCabe, E.~E.; and Edgerton, E. 2019.
\newblock Tweeting for social justice in \#Ferguson: Affective discourse in
  Twitter hashtags.

\bibitem[{Carney(2016)}]{doi:10.1177/0160597616643868}
Carney, N. 2016.
\newblock All Lives Matter, but so Does Race: Black Lives Matter and the
  Evolving Role of Social Media.
\newblock \emph{Humanity \& Society} 40(2): 180--199.
\newblock \doi{10.1177/0160597616643868}.

\bibitem[{Dacon and Liu(2021)}]{10.1145/3442442.3452325}
Dacon, J.; and Liu, H. 2021.
\newblock \emph{Does Gender Matter in the News? Detecting and Examining Gender
  Bias in News Articles}, 385–392.
\newblock New York, NY, USA: Association for Computing Machinery.
\newblock ISBN 9781450383134.

\bibitem[{Freelon, McIlwain, and Clark(2016)}]{Freelon2016BeyondTH}
Freelon, D.; McIlwain, C.; and Clark, M. 2016.
\newblock Beyond the Hashtags: \#Ferguson, \#Blacklivesmatter, and the Online
  Struggle for Offline Justice.
\newblock \emph{Social Science Research Network} .

\bibitem[{Gallagher et~al.(2017)Gallagher, Reagan, Danforth, and
  Dodds}]{gallagher2017divergent}
Gallagher, R.~J.; Reagan, A.~J.; Danforth, C.~M.; and Dodds, P.~S. 2017.
\newblock Divergent discourse between protests and counter-protests:
  \#BlackLivesMatter and \#AllLivesMatter.

\bibitem[{Garza(2014)}]{Garza:2014}
Garza, A. 2014.
\newblock A Herstory of the \#BlackLivesMatter Movement by Alicia Garza.

\bibitem[{Giorgi et~al.(2020)Giorgi, Guntuku, Rahman, Himelein-Wachowiak,
  Kwarteng, and Curtis}]{giorgi2020twitter}
Giorgi, S.; Guntuku, S.~C.; Rahman, M.; Himelein-Wachowiak, M.; Kwarteng, A.;
  and Curtis, B. 2020.
\newblock Twitter Corpus of the \#BlackLivesMatter Movement And Counter
  Protests: 2013 to 2020.

\bibitem[{Hamborg, Donnay, and Gipp(2018)}]{Hamborg2018AutomatedIO}
Hamborg, F.; Donnay, K.; and Gipp, B. 2018.
\newblock Automated identification of media bias in news articles: an
  interdisciplinary literature review.
\newblock \emph{International Journal on Digital Libraries} 1--25.

\bibitem[{Ince, Rojas, and Davis(2017)}]{ince2017response}
Ince, J.; Rojas, F.; and Davis, C.~A. 2017.
\newblock The social media response to Black Lives Matter: how Twitter users
  interact with Black Lives Matter through hashtag use.
\newblock \emph{Ethnic and Racial Studies} .

\bibitem[{Jackson(2020)}]{Jackson2020OnA}
Jackson, S. 2020.
\newblock On \#BlackLivesMatter and Journalism.
\newblock \emph{Sociologia} 14: 101--108.

\bibitem[{Keib, Himelboim, and Han(2018)}]{KEIB2018106}
Keib, K.; Himelboim, I.; and Han, J.-Y. 2018.
\newblock Important tweets matter: Predicting retweets in the
  \#BlackLivesMatter talk on twitter.
\newblock \emph{Computers in Human Behavior} .

\bibitem[{Mundt, Ross, and Burnett(2018)}]{Mundt2018ScalingSM}
Mundt, M.; Ross, K.; and Burnett, C.~M. 2018.
\newblock Scaling Social Movements Through Social Media: The Case of Black
  Lives Matter.
\newblock \emph{Social Media + Society} .

\bibitem[{Nowell et~al.(2017)Nowell, Norris, White, and
  Moules}]{doi:10.1177/1609406917733847}
Nowell, L.~S.; Norris, J.~M.; White, D.~E.; and Moules, N.~J. 2017.
\newblock Thematic Analysis: Striving to Meet the Trustworthiness Criteria.
\newblock \emph{International Journal of Qualitative Methods} 16(1):
  1609406917733847.
\newblock \doi{10.1177/1609406917733847}.

\bibitem[{Reissner and Çalışkan(2019)}]{Reissner2019FromT}
Reissner, K.; and Çalışkan, G. 2019.
\newblock From \#BlackLivesMatter to Black Liberation.
\newblock \emph{Ethnic and Racial Studies} 42: 1394 -- 1396.

\bibitem[{Rezapour(2018)}]{rezapour2018using}
Rezapour, R. 2018.
\newblock Using Linguistic Cues for Analyzing Social Movements.

\bibitem[{Tsur and Rappoport(2012)}]{10.1145/2124295.2124320}
Tsur, O.; and Rappoport, A. 2012.
\newblock What's in a Hashtag? Content Based Prediction of the Spread of Ideas
  in Microblogging Communities.
\newblock New York, NY, USA: Association for Computing Machinery.
\newblock ISBN 9781450307475.
\newblock \doi{10.1145/2124295.2124320}.

\bibitem[{Williams(1975)}]{Williams}
Williams, A. 1975.
\newblock Unbiased Study of Television News Bias.
\newblock \emph{Journal of Communication} .

\bibitem[{Yang(2016)}]{Yang2016NarrativeAI}
Yang, G. 2016.
\newblock Narrative Agency in Hashtag Activism: The Case of \#BlackLivesMatter.
\newblock \emph{Media and Communication} .

\bibitem[{Zulli(2020)}]{Zulli2020EvaluatingHA}
Zulli, D. 2020.
\newblock Evaluating hashtag activism: Examining the theoretical challenges and
  opportunities of \#BlackLivesMatter.

\end{thebibliography}

\end{document}